\documentclass[lettersize,journal]{IEEEtran}
\usepackage{amsmath,amsfonts}
\usepackage{algorithm}
\usepackage{array}
\usepackage[caption=false,font=normalsize,labelfont=sf,textfont=sf]{subfig}
\usepackage{textcomp}
\usepackage{stfloats}
\usepackage{url}
\usepackage{verbatim}
\usepackage{graphicx}
\usepackage{cite}

\usepackage{makecell}
\usepackage{algpseudocode}
\usepackage{cleveref}
\usepackage{tabularx}
\usepackage{booktabs}
\usepackage{float}
\usepackage{multicol}
\usepackage{multirow}
\usepackage{amssymb}
\usepackage{pdflscape}

\algrenewcommand\algorithmicrequire{\textbf{Input:}}
\algrenewcommand\algorithmicensure{\textbf{Output:}}
\begin{document}

\title{Randomized Histogram Matching: A Simple Augmentation for Unsupervised Domain Adaptation in Overhead Imagery}

\author{Can Yaras, Kaleb Kassaw, Bohao Huang, Kyle Bradbury, Jordan M. Malof
        % <-this % stops a space
\thanks{Can Yaras is with the Department of Electrical and Computer Engineering at University of Michigan, Ann Arbor, MI 48109. (email: cjyaras@umich.edu)}% <-this % stops a space
\thanks{Kaleb Kassaw and Bohao Huang are with the Department of Electrical and Computer Engineering at Duke University, Durham, NC 27708. (email: kaleb.kassaw@duke.edu, bohao.huang@duke.edu)}% <-this % stops a space
\thanks{Kyle Bradbury is with the Nicholas Institute for Energy, Environment, and Sustainability at Duke University, Durham, NC 27708. (email: kyle.bradbury@duke.edu)}% <-this % stops a space
\thanks{Jordan M. Malof is with the Department of Computer Science at the University of Montana, Missoula, MT 59812. (email: jordan.malof@umt.edu)}% <-this % stops a space
\thanks{Manuscript received Month Day, 2023; revised Month Day, 2023.}}

% The paper headers
\markboth{IEEE Journal of Selected Topics in Applied Earth Observations and Remote Sensing
,~Vol.~??, No.~??, ??August~??2021}%
{Shell \MakeLowercase{\textit{Yaras et al.}}: Randomized Histogram Matching: A Simple Augmentation for Unsupervised Domain Adaptation in Overhead Imagery}

\IEEEpubid{0000--0000/00\$00.00~\copyright~2021 IEEE}
% Remember, if you use this you must call \IEEEpubidadjcol in the second
% column for its text to clear the IEEEpubid mark.

\maketitle

\begin{abstract}
Modern deep neural networks (DNNs) are highly accurate on many recognition tasks for overhead (e.g., satellite) imagery. However, visual domain shifts (e.g., statistical changes due to geography, sensor, or atmospheric conditions) remain a challenge, causing the accuracy of DNNs to degrade substantially and unpredictably when testing on new sets of imagery. In this work, we model domain shifts caused by variations in imaging hardware, lighting, and other conditions as non-linear pixel-wise transformations, and we perform a systematic study indicating that modern DNNs can become largely robust to these types of transformations, if provided with appropriate training data augmentation. In general, however, we do not know the transformation between two sets of imagery. To overcome this, we propose a fast real-time unsupervised training augmentation technique, termed randomized histogram matching (RHM). We conduct experiments with two large benchmark datasets for building segmentation and find that despite its simplicity, RHM consistently yields similar or superior performance compared to state-of-the-art unsupervised domain adaptation approaches, while being significantly simpler and more computationally efficient. RHM also offers substantially better performance than other comparably simple approaches that are widely used for overhead imagery.
\end{abstract}


\section{Introduction}
\IEEEPARstart{M}{odern} deep neural networks (DNNs) can now achieve accurate recognition on a variety of tasks involving overhead imagery (e.g., satellite imagery, aerial photography), such as classification, object detection, and semantic segmentation \cite{demir2018deepglobe,van2018spacenet,sergievskiy2019reduced}. One emergent limitation of DNNs in remote sensing, however, is their sensitivity to the statistics of their training imagery. Recent research has shown that DNNs often perform unpredictably, and often much more poorly, when they are applied to novel collections with respect to their training data \cite{tasar2020colormapgan,kong2020synthinel,maggiori2017can,huang2020deep}. Furthermore, this performance degradation seems to occur even if DNNs are trained on relatively large and diverse datasets, encompassing large and diverse geographic regions \cite{kong2020synthinel,huang2020deep}.

\begin{figure}[ht]
\centering
\includegraphics[width=0.7\linewidth]{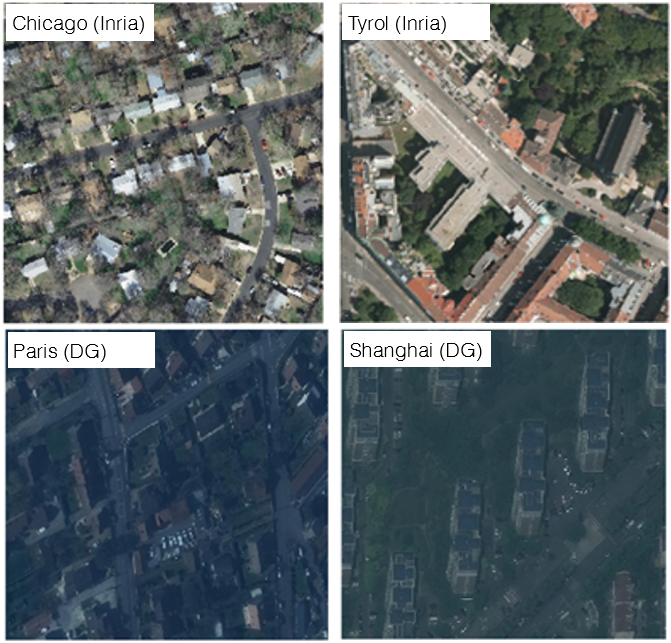}
\caption{An illustration of the domain shifts between different collections of overhead imagery. These are representative images from two cities from the Inria and DeepGlobe (DG) datasets.  Both Inria and DG serve as our experimental datasets in this work.}
\label{fig:domain_shift_visual_examples}
\end{figure}

One cause of the performance degradation of DNNs on new sets of imagery involves visual domain shift (i.e., distribution shift); these are statistical differences between the training imagery and new collections of imagery \cite{tasar2020colormapgan,kong2020synthinel}. \Cref{fig:domain_shift_visual_examples} presents images from different collections of imagery where the domain shift is readily visible. These domain shifts are caused by variations in a diverse set of factors that influence the appearance (i.e., statistics) of the overhead imagery including scene geography, the built environment (e.g., building and road styles), imaging hardware, weather, time-of-day, and others. Each of these factors influences the imagery in a manner that is generally complex and unknown in advance, and therefore challenging to address.

One straightforward solution to these domain shifts is to label a subset of each new collection of imagery and then re-train the DNN; however, this solution is costly and time-consuming \cite{tasar2020colormapgan,kong2020synthinel}. Instead, we would ideally have a model that performs well across many different collections of imagery and does so without the need for labels from each one.  This setting is a special case of a broader problem in machine learning known as \textit{unsupervised domain adaptation}, wherein it is assumed that we are given a ``source domain'' dataset with ground truth labels and that we aim to maximize recognition performance on one (or more) sets of unlabeled ``target domain'' data \cite{tuia2016domain}.

\IEEEpubidadjcol

\subsection{Spectral domain shift and adaptation}
The unsupervised domain adaptation problem has been studied extensively in recent years \cite{wang2018deep,patel2015visual}, and has also recently received growing attention in the remote sensing community due to the aforementioned challenges of domain shifts \cite{tuia2016domain,tasar2020colormapgan,tasar2020standardgan,tasar2020semi2i,tasar2020daugnet}. Most recent domain adaptation approaches for overhead imagery attempt to address all sources of domain shift simultaneously.  In this work, however, we attempt to simplify the problem by focusing on a subset of domain shifts that can be modeled as purely spectral (single-pixel) transformations. Mathematically, this is given by
\begin{equation}\label{eq:normalization_shift}
    p^{t} = T(p^{s})
\end{equation}
where $p^{s}$ and $p^{t}$ are the source and target domain pixel intensities, respectively, of an otherwise identical scene.  We hypothesize that domain shifts of this type arise from variations across imagery collections in several specific factors: e.g., camera specifications and calibration, time of day, and lighting conditions.  Variations in these factors are likely to occur, to varying degrees, between almost any two collections of imagery so that domain shifts of the kind in \eqref{eq:normalization_shift} are common.  Our experiments here suggest that this is not only the case, but that spectral domain shifts appear to be responsible for a significant proportion of the performance degradation of DNNs when applied to novel collections of imagery. 

\subsection{Contributions of this work}\label{subsec:contributions_of_this_work}

In this work, we begin by investigating whether spectral domain shifts of the kind in  \eqref{eq:normalization_shift} can be addressed simply through data augmentation during training.  We perform a systematic study where we train DNNs of varying capacity (i.e., number of free parameters) with image augmentation comprising different classes of spectral augmentations (e.g., gamma, affine, etc).  We then test the performance of these networks on collections of imagery that have been augmented with one of these same classes of spectral transformations.  We find that modern DNNs with large encoders (e.g., ResNet-18, 50, 100 \cite{he2016deep}) can become largely robust to several different classes of spectral transformations if provided with a \textit{matching} training augmentation strategy. In general, however, we do not know the transformation between any two collections of imagery, or even the class of transformations from which it may be drawn (e.g., affine, gamma), so it is unclear which augmentation should be adopted. 

\begin{figure}[H]
\centering
\includegraphics[width=0.95\linewidth]{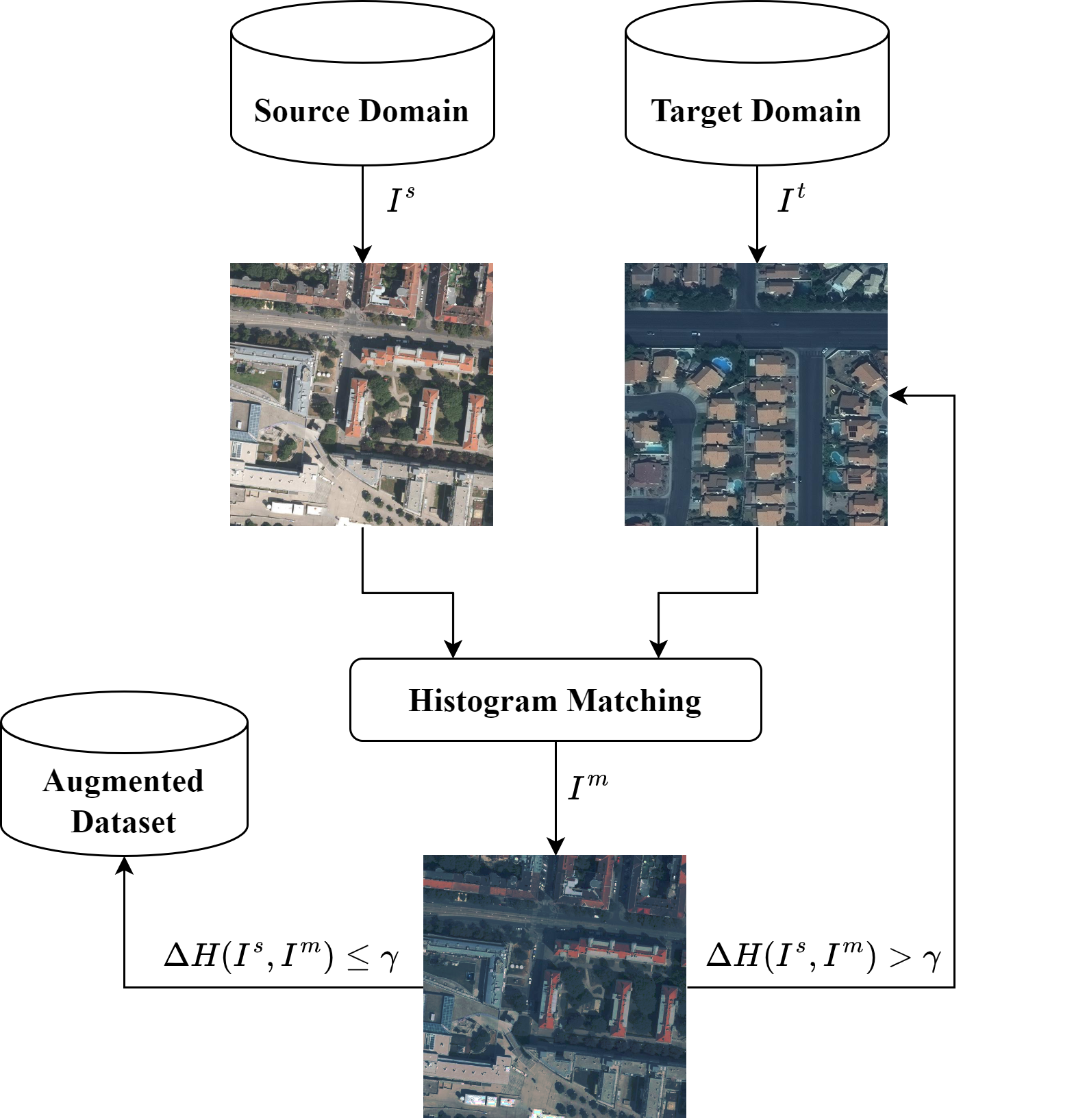}
\caption{An illustration of the randomized histogram matching concept. To produce an augmented training dataset, we repeat the following process. For each image $I^s$ in the training dataset (source domain), an image $I^t$ is drawn randomly from the testing dataset (unlabeled target domain). Then the histogram of $I^s$ is matched to the histogram of $I^t$, which yields the modified image $I^m$. If the information loss $\Delta H$ (defined in \Cref{subsec:entropy_based_resampling}) between $I^m$ and $I^s$ is below a set threshold $\gamma$, $I^m$ is added to the augmented training dataset. Otherwise, another random image is drawn from the target domain and the process is repeated. This resampling is only performed at most once.}
\label{fig:rhm_diagram}
\end{figure}

To overcome this problem, we propose a simple augmentation technique, termed \textit{randomized histogram matching} (RHM), that matches the histogram of each training image to a randomly-chosen (unlabeled) target domain image, as illustrated in \Cref{fig:rhm_diagram}. This approach results in a random spectral shift being applied to \textit{each} training image, and we hypothesize that this occasionally (by chance) approximates the true spectral shift between the source and target domains (see Sec. \ref{sec:rhm}. Consequently, the training data are occasionally augmented (again, by chance) with the true spectral shift and can thereby become more robust to it. Since RHM only requires the unlabeled testing data, it can be viewed as a simple unsupervised domain adaptation approach. 

To demonstrate the efficacy of RHM, we conduct benchmark testing with two large publicly-available datasets for building segmentation in two settings: (1) training on one collection and testing on one collection (one-to-one adaptation, following \cite{tasar2020daugnet}), and (2) a more real-world scenario where we train on multiple domains and test on multiple domains (many-to-many, also following \cite{tasar2020daugnet}).  We focus on building segmentation because it is a challenging task that has received substantial attention in recent years, and there are large and diverse benchmark datasets to support our multi-domain benchmark experiments (e.g., Inria \cite{maggiori2017can} and DG \cite{demir2018deepglobe}). We now summarize our contributions as follows: 

\begin{itemize}

\item \textit{Can augmentation confer spectral robustness in DNNs?} We provide, to our knowledge, the first systematic empirical evidence that modern DNNs with large encoders are capable of becoming robust to complex spectral transformations via data augmentation during training.  We show that the ability of DNNs to become robust depends upon their capacity (i.e., number of free parameters), especially for more complex classes of transformations.  We also find that DNNs only become robust to the precise class of spectral transforms that were used for augmentation, rather than becoming robust to generic spectral transforms.  This suggests that augmentation is an effective mechanism to address spectral domain shifts, if a class of spectral transforms is used that includes in it the particular transforms that are often encountered in real-world imagery (e.g., between independent collections of imagery).    

\item \textit{Randomized Histogram Matching (RHM) augmentation}. RHM is a simple, yet highly effective unsupervised domain adaptation approach, via spectral augmentation. We show that RHM almost always offers substantial performance benefits in unsupervised cross-domain settings (e.g., we wish to apply a model in a new geo-location with no labeled data). Our results also indicate RHM usually offers substantially greater performance benefits than other common types of spectral augmentation  (e.g., Affine, Gamma, or HSV), and it also performs competitively with (or even better than) two state-of-the-art unsupervised domain adaptation approaches (i.e., CycleGAN or ColorMapGAN \cite{tasar2020colormapgan}), despite being substantially simpler and faster in many cases (e.g., RHM only has one hyperparameter, and does not require training additional models).

\end{itemize}

Next, in \Cref{sec:related_work}, we provide further details about related work and how our contributions differ from them.  

\section{Related Work}\label{sec:related_work}
In this section, we review related work for unsupervised domain adaptation in overhead imagery and how our work relates and builds upon it.  Unsupervised adaptation methods can be divided into two broad groups: model adaptation and data adaptation.  

\subsection{Unsupervised model adaptation}  In most of these approaches, the goal is to obtain features (e.g., through selection or learning) that are invariant across the source and target domains, but still useful to discriminate between the target classes.  Some approaches have focused on a feature selection strategy \cite{bruzzone2009domain,persello2015kernel}.  Most other approaches attempt to learn the desired feature representation.  A large variety of approaches have been proposed using this underlying strategy (e.g., \cite{huang2018domain,tsai2018learning,hoffman2016fcns}, including for remote sensing imagery (e.g., \cite{deng2019large,zhang2018fully,zou2018unsupervised,bruzzone2001unsupervised,bruzzone2002partially,qin2019tensor}).   

\subsection{Unsupervised data adaptation}  Our work here builds directly upon recently proposed methods of this kind.  These methods are designed to modify the source and/or target domain data so that they are statistically more similar to each other.  If successful, a recognition model that is trained and evaluated on the modified source and target data should be more accurate.  These methods can be subdivided into two main categories: (i) domain standardization and (ii) domain matching.  In (i) the goal is to map the source and target domains into some common domain.  Some well-known examples of such approaches are normalization (or z-scoring) \cite{gonzalez2002digital}; histogram equalization \cite{gonzalez2002digital}, color invariance approaches (e.g., \cite{buchsbaum1980spatial,pacifici2014importance,forsyth1990novel,itten1993geometric}), and recent approaches using DNNs \cite{tasar2020standardgan}.   

In (ii) the goal is to match the source domain to the target domain.  Graph matching \cite{tuia2012graph,das2018unsupervised} and especially histogram matching \cite{tasar2020standardgan} are common approaches for this.  Based on the CycleGAN model \cite{zhu2017unpaired}, a large number of approaches have been proposed to train a DNN to map source domain data to be more similar to the target domain  \cite{hoffman2018cycada,liu2017unsupervised,huang2018multimodal,lee2018diverse,benjdira2019unsupervised,tasar2020colormapgan}.  One challenge with many of these approaches is that they can alter the semantic content of the source domain imagery \cite{tasar2020colormapgan} (e.g., changing object shapes or even their semantic class).  

More recently, ColorMapGAN \cite{tasar2020colormapgan} was proposed to address this challenge by restricting the DNN to perform pixel-wise intensity transformations, preventing the model from making more complex semantic changes to the imagery. The authors show that ColorMapGAN (along with CycleGAN \cite{zhu2017unpaired}) outperformed a variety of types of unsupervised domain adaptation approaches for segmentation on overhead imagery.  One limitation of ColorMapGAN, however, is that a separate model has to be learned between each \textit{pair} of source and target domains, which is impractical for a large number of source and target domains (many-to-many testing). In this work we propose RHM as a simple and fast alternative to recent DNN-based unsupervised domain adaptation approaches.

\subsection{Data augmentation} 
In this approach, the original training dataset is supplemented with transformed yet semantically consistent views of the training data that create variations in the training imagery. Some commonly utilized classes of transformations used for augmentation in remote sensing are Gamma corrections \cite{buslaev2020albumentations} and contrast changes (e.g., via Hue-Saturation-Value (HSV) shifts \cite{buslaev2020albumentations}).  These approaches are designed to build invariance to different classes of spectral shift.  For this reason, we will compare RHM to Gamma and HSV augmentation approaches, and investigate whether DNNs can indeed become robust to these transformations, as is implicitly assumed in their application. 

\section{Experimental Datasets}\label{sec:experimental_dataset}

In our experiments we employ two large publicly-available datasets for building segmentation: Inria \cite{maggiori2017can} and DeepGlobe (termed ``DG'') \cite{demir2018deepglobe}. Both datasets are composed of high-resolution (0.3m) color overhead imagery and have accompanying pixel-wise building labels. Both datasets include large quantities of imagery from several distant geographic locations, summarized in \Cref{tab:datasets}.  Importantly, each collection varies greatly in both their scene content and their spectral characteristics -- \Cref{fig:domain_shift_visual_examples} presents examples of imagery from DG and Inria illustrating these differences. 

\begin{table}[h] 
\captionsetup{justification=centering}
\caption{Cities included in Inria and DeepGlobe with corresponding surface area of imagery.}
\label{tab:datasets}
\centering
\newcolumntype{C}{>{\centering\arraybackslash}X}
\begin{tabularx}{\linewidth}{CCC}
\toprule
\textbf{Dataset} & \textbf{City} & \textbf{Surface Area (km$^2$)} \\
\midrule
\multirow{5}{*}{Inria \cite{maggiori2017can}} & Austin, USA & 81 \\
& Chicago, USA & 81 \\
& Kitsap, USA & 81 \\
& West Tyrol, Austria & 81 \\
& Vienna, Austria & 81 \\
\midrule
\multirow{4}{*}{DeepGlobe \cite{demir2018deepglobe}} & Las Vegas, USA & 113 \\
& Paris, France & 33 \\
& Shanghai, China & 133 \\
& Khartoum, Sudan & 29 \\
\bottomrule
\end{tabularx}
\end{table}

\section{Segmentation Model and Training}\label{sec:segmentation_model}

In recent years, U-Net \cite{ronneberger2015u} and its variants (e.g., \cite{iglovikov2018ternausnet}) have achieved state-of-the-art performance for building segmentation in overhead imagery (e.g., \cite{demir2018deepglobe,maggiori2017can}). Following \cite{iglovikov2018ternausnet}, we modify U-Net by using ResNet encoders of varying size that have been pretrained on the ImageNet dataset.  In our experiments we will use a ResNet-50 encoder unless otherwise noted, to balance training speed and performance.  We also make the following specific design choices for our models: (a) cross-entropy loss between the pixel-wise ground truth and predictions, (b) the SGD optimizer, (c) 90 epochs of training, and (d) a batch size of 8.  We also use a learning rate of 0.001 and 0.01, respectively, for the encoder and decoder of the U-Net models.  A smaller learning rate is applied to the encoder since it is already pretrained on ImageNet. For both the encoder and decoder, we drop the learning rate by one order of magnitude after 50 and 80 epochs. These settings are chosen to be nearly identical to that those in \cite{nair2020designing} -- the only variation is that we additionally drop the learning rate after 80 epochs to ensure that the validation loss converges by the end of training.%

\section{Baseline Adaptation Methods}\label{sec:baseline_methods}
For baselines, we focused upon spectral adaptation methods, allowing us to compare RHM to methods with similar complexity (e.g., only altering spectral content of the imagery).  We include computationally simple spectral augmentations that are widely-used in remote sensing (e.g., Gamma, HSV, Affine intensity augmentations), as well as ColorMapGAN, a recent \textit{data-driven} spectral adaptation method.  We also consider one state-of-the-art method that is not restricted to spectral transforms, CycleGAN, to determine how spectral methods compare with more general adaptation methods.  

\subsection{Augmentation}
We consider three parameterized transformations as baseline augmentations for comparison to RHM: \textit{Affine}, \textit{Gamma}, and \textit{HSV}. These are common transformations for modeling spectral transformations and as such are most relevant for comparison to our proposed method. \Cref{tab:augmentations} contains the parameterized functional forms and their respective distributions. The distribution of each parameter is chosen via a standard Bayesian hyperparameter optimization procedure using Gaussian processes, as described in \cite{snoek2012practical}. For the HSV augmentation, we fix the scaling factor of the hue channel $\alpha^{(H)}$ to be 1, since the hue value corresponds to the angular dimension in the cylindrical geometry of HSV space. During training, we apply these augmentations in real time throughout training, with uniquely sampled augmentation parameters for each mini-batch iteration. 

\begin{table}[h]
\centering
\captionsetup{justification=centering}
\caption{Parameterized forms of transformations for each baseline augmentation.}
\label{tab:augmentations}
%\newcolumntype{C}{>{\centering\arraybackslash}X}
\begin{tabularx}{\linewidth}{ccc}%{CCC}
\toprule
\textbf{Augmentation} & \textbf{Transformation} & \textbf{Parameters} \\
\midrule
 Affine & \makecell{$X^{(c)} \mapsto \alpha^{(c)} X^{(c)} + \mu^{(c)}$ \\ $c \in \{R, G, B\}$} & \makecell{$\alpha^{(c)}  \stackrel{iid}{\sim} \mathcal{U}(0.82, 1.18)$ \\ $\mu^{(c)} \stackrel{iid}{\sim} \mathcal{U}(-0.38, 0.38)$} \\ 
 \midrule
 Gamma & \makecell{$X^{(c)} \mapsto \left(X^{(c)}\right)^{\gamma^{(c)}}$ \\ $c \in \{R, G, B\}$} & $\gamma^{(c)} \stackrel{iid}{\sim} \mathcal{U}(0.32, 1.68)$ \\
 \midrule
 HSV & \makecell{$Y^{c} \mapsto \alpha^{(c)} Y^{(c)} + \mu^{(c)}$ \\ $Y = HSV(X)$ \\ $c \in \{H, S, V\}$} & \makecell{$\alpha^{(S)}, \alpha^{(V)} \stackrel{iid}{\sim} \mathcal{U}(0.63, 1.37)$ \\ $\alpha^{(H)} = 1$ \\ $\mu^{(c)} \stackrel{iid}{\sim} \mathcal{U}(-0.27, 0.27)$} \\
\bottomrule
\end{tabularx}
\end{table}

\subsection{Standardization}

\subsubsection{Histogram Equalization \cite{gonzalez2002digital}} One approach to standardizing each domain is to ensure that the contrast of all images are the same. Histogram equalization achieves this by adjusting the histogram of pixel intensities of each image to be uniform. We transform images in both the source and target domain in this manner and use the transformed images for training and testing the U-Net model, respectively.

\subsubsection{Gray World \cite{buchsbaum1980spatial}} This approach attempts to find a standardized domain in which each image's average color is gray, and therefore invariant to illumination conditions that may affect each color channel independently. By modeling the deviation in color illumination of each channel from gray as a linear scaling, we may remove the scaling factor by normalizing each channel by its average pixel intensity. We transform images in both the source and target domain in this manner and use the transformed images for training and testing of the U-Net model, respectively.

\subsection{Image-to-image translation}

\subsubsection{Histogram Matching \cite{gonzalez2002digital}} A naive method for matching the distribution of the source domain to the target domain is to match the histogram of each source image to the aggregate histogram of the target domain. This matching is done independently for each channel. We transform only the images in the source domain in this manner and use the transformed images for training the U-Net model.

\subsubsection{ColorMapGAN \cite{tasar2020colormapgan}} This state-of-the-art approach aims to learn an unconstrained pixel-wise mapping from the source to target domain, modeled as a generator in an unsupervised adversarial setting. As with most GAN set-ups, there is a generator $G$ and a discriminator $D$, where $D$ attempts to differentiate images generated by $G$ from the images in the target domain. On the other hand, $G$ learns a unique pixel-to-pixel mapping for every possible RGB triple. $G$ and $D$ are trained simultaneously with the LSGAN \cite{mao2017lsgan} loss. After training, $G$ is used to generate fake images from the source domain that look like the target domain - these fake source images are then used to train the U-Net model. We use the same hyperparameters as given in \cite{tasar2020colormapgan} in our own experiments.

\subsubsection{CycleGAN \cite{zhu2017unpaired}} Similar to ColorMapGAN, this method learns a transformation between domains in an unsupervised adversarial setting. However, we now have two generators $G$ and $F$ where $G$ attempts to transform the source domain $S$ to the target domain $T$ and $F$ attempts to transform the target domain $T$ to the source domain $S$. Unlike ColorMapGAN, both $G$ and $F$ are multi-layer networks that can realize more complicated functions than pixel-wise transforms. We also have two domain-specific discriminators $D_S$ and $D_T$ that attempt to differentiate the real and fake images in their respective domains. $G$ and $D$ are trained simultaneously via an objective that combines the conditional GAN \cite{mirza2014conditional} loss in both directions with a cycle-consistency loss term, which forces the compositions $F \circ G$ and $G \circ F$ to be the identity mapping. After training, $G$ is used to generate fake images from the source domain that look like the target domain - these fake source images are then used to train the U-Net model. We use the same hyperparameters as given in \cite{zhu2017unpaired} in our own experiments.

\section{Randomized Histogram Matching (RHM) Augmentation} \label{sec:rhm}
RHM is a modification of conventional histogram matching (HM).  In conventional HM, one matches the histogram of pixel intensities of one set of imagery to the corresponding histogram of another set of imagery.  For example, for cross-domain adaptation, we can transform a single source image to match the histogram of created from the full collection of target domain imagery (see Sec. \ref{sec:baseline_methods}). This approach works well if the scene content of the source imagery and the target imagery are similar, in which case the histograms of the two image collections \textit{should} be similar as well; any differences must be due to other factors (e.g., variations in lighting, imaging hardware, etc), which are often modeled as spectral domain shifts. Consequently, matching the histograms removes any existing spectral domain shift between the two image sets, if their scene content is similar. However, we hypothesize that it is unlikely that two random sets of imagery will contain similar scene content.  In such cases, the differences in histograms will be due to content differences, in which case the two histograms will not generally be the same, even if other imaging conditions are similar (e.g., lighting, hardware).  When such content differences are present, therefore, conventional HM produces undesirable results.  For example, our experimental results in Sec. \ref{sec:one_to_one} indicate that conventional HM often works well, but that it is also inconsistent, and can sometimes fail badly. 

RHM is intended to mitigate this limitation of HM by relying upon matching many random pairs of imagery, making it likely that the content between the pairs will sometimes (by chance) be similar, causing HM to approximate the true underlying spectral shift for those pairs.  Specifically, as outlined in Fig. \ref{fig:rhm_diagram}, RHM matches the histogram of each training image with the histogram of a randomly-sampled target domain image.  We hypothesize that this approach often results in image pairs with dissimilar scene content (like conventional HM), but that it also sometimes creates pairs with similar scene content.  Consequently, some training images are augmented in a way that approximates the true underlying spectral shift between the source and target domain imagery.  Furthermore, we hypothesize that largely inaccurate augmentations will be effective for training robust DNNs as long as it \textit{periodically} augments with the correct transformations. This is motivated by the experiments in Sec. \ref{sec:does_augmentation_confer_invariance} suggesting that a given class of spectral augmentations will work well as long as the true augmentation is a special case of the class (e.g., augmentation with \textit{random} spectral Affine transforms works well for a given target-domain if that target domain is shifted by \textit{any specific} Affine transform).

\subsection{The RHM Algorithm}
The detailed RHM training procedure is summarized in \Cref{alg:rhm}, and we describe two of the major components in more detail:  histogram matching, and entropy-based resampling. 

\textbf{Histogram matching.} Mathematically, the histogram matching is performed as follows: for a source image $X \in \mathbb{R}^{C \times H \times W}$, let $F_c: \mathbb{R} \rightarrow [0, 1]$ be the normalized cumulative histogram of each channel $c \in [C]$, i.e., $F_c(x)$ is the proportion of pixels in channel $c$ with magnitude no more than $x$. Similarly, define $G_c$ to be the normalized cumulative histogram of channel $c \in [C]$ for a randomly selected target image $\widetilde{X} \in \mathbb{R}^{C \times H' \times W'}$. Then the RHM augmented version $T(X; \widetilde X)$ of source $X$ with target $\widetilde X$ is defined as
\begin{equation}\label{eq:histogram_matching}
    T(X; \widetilde X)_{c, i, j} = G_c^{-1}(F_c(X_{c, i, j}))
\end{equation}
where $G_c^{-1}(y) \triangleq \min \{x: G_c(x) \geq y\}$.

\begin{figure}[h]
\centering
\includegraphics[width=\linewidth]{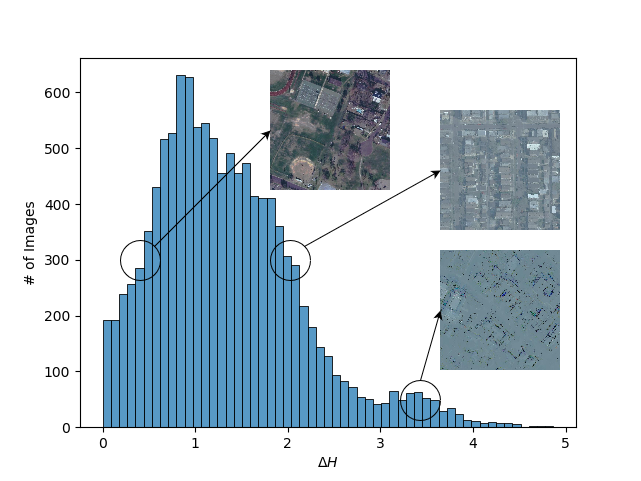}
\caption{Distribution of changes in image information $\Delta H$ resulting from RHM transforms. Higher values of $\Delta H$ correspond to greater degrees of loss in image information.}
\label{fig:d_entropy_hist_w_images}
\end{figure}

\textbf{Entropy-based resampling.}\label{subsec:entropy_based_resampling} For some pairings of source and target images, the transform in \eqref{eq:histogram_matching} can result in a large loss of image information needed to perform the building segmentation task, which can be detrimental to model training.  To limit the amount of image compression in RHM augmentations, we discard augmentations that lead to large compressions of the image intensity values.  We measure the compression via the change in Shannon entropy of their histograms, here denoted $H(X)$ for an image $X$ and $H(T(X))$ for a transformed image, where $H(X)$ is defined as
\begin{equation}\label{image_entropy}
    % H(X) = -\frac{1}{c}\sum_{c \in [C]} \sum_{i=0}^{255}\Pr[f_c = i]\log(\Pr[f_c = i]).
    H(X) = -\frac{1}{c}\sum_{c \in [C]} \int  f_c(x) \log(f_c(x))\, dx
\end{equation}
where $f_c$ is the normalized histogram of each channel $c \in [C]$ of $X$. We then define the quantity $\Delta H$, the change in image information, as
\begin{equation}
    \Delta H \triangleq H(X) - H(T(X)).
\end{equation}
The distribution of values of $\Delta H$ for RHM transforms is shown in \Cref{fig:d_entropy_hist_w_images} along with the associated augmented images. Here the loss of image information is apparent as $\Delta H$ increases. 

By evaluating the value of $\Delta H$ for each source-target image pair during training, and excluding those pairs with high $\Delta H$,  we limit the resulting image compression from RHM transforms. We set a threshold $\gamma$, and in cases where RHM transforms produce $\Delta H > \gamma$, the first pairing is discarded, a new randomly selected target image is chosen, and the transformation in \eqref{eq:histogram_matching} is computed once more. To limit overall computation time, we only repeat this resampling step once for each source image.  We show in our experiments in \Cref{sec:one_to_one} and \Cref{sec:many_to_many} that this filtering step is consistently beneficial.

\begin{algorithm}
\caption{Randomized Histogram Matching}\label{alg:rhm}
\begin{algorithmic}
\Require Labeled source domain $\mathcal D_S = \{(X_i, Y_i)\}$, unlabeled target domain $\mathcal D_T = \{\widetilde{X}_j\}$, training epochs $T$, batch size $B$, learning rate schedule $(\eta_t)_{t\geq 1}$, segmentation loss $\ell(\cdot, \cdot)$
\Ensure Segmentation model $f_{\theta^*}(\cdot)$ adapted to $\mathcal D_T$
\State Initialize (pretrained) segmentation model $f_{\theta}(\cdot)$
\For{epoch $t = 1, \dots, T$}
    \State $\mathcal{B}_t \gets $ shuffle and group $\mathcal D_S$ into batches of size $B$
    \For{batch $\{(X_i, Y_i)\}_{i=1}^B$ in $\mathcal{B}_t$}
        \For{$i = 1, \dots, B$}
            \State $\widetilde X_i \gets $ uniformly sampled from $\mathcal D_T$
            \State $Z_i \gets T(X_i; \widetilde X_i)$
            \If{$H(X_i) - H(Z_i) > \gamma$}
                \State $\widetilde X_i \gets $ uniformly sampled from $\mathcal D_T$
                \State $Z_i \gets T(X_i; \widetilde X_i)$
            \EndIf
        \EndFor
        \State $\theta \gets \theta - \eta_t \nabla_{\theta} \sum_i \ell(f_{\theta}(Z_i), Y_i)$
    \EndFor
\EndFor
\State $\theta^* \gets \theta$
\end{algorithmic}
\end{algorithm}

As a result, we are able to utilize variations in the target domain to apply random spectral shifts to the training imagery that we hypothesize periodically coincide with the true spectral shift between the source and target domains. Like the baseline augmentations described in \Cref{sec:baseline_methods}, we apply RHM as an online augmentation to each training image, where a new target image is sampled every iteration for matching. See \Cref{alg:rhm} for a full description of training using RHM with entropy resampling. RHM is not applied to the target domain during testing.

\section{Can Augmentation Confer Robustness to Spectral Transformations?}
\label{sec:does_augmentation_confer_invariance}

In this section, we investigate the extent to which modern DNNs with high-capacity feature encoders can become robust to spectral transformations of the form in \eqref{eq:normalization_shift}. Many popular augmentation approaches for remote sensing (and elsewhere) apply random intensity transformations (e.g., HSV, Gamma) with the implicit assumption that DNNs will become robust to domain shifts with a similar functional form. To our knowledge, there have been no controlled experiments investigating these assumptions or evaluating how they depend either on the complexity of the spectral transformations, or the capacity of the DNNs involved.  We also investigate whether augmentation with one class of spectral functions (e.g., HSV) confers general invariance to spectral transformations (e.g., the DNN learns to ignore spectral shifts of any kind), or that the DNN only becomes robust to the particular class of functions it was trained upon.  

To address these questions, we evaluate the performance of state-of-the-art segmentation models on test datasets that exhibit five different kinds of augmentations (i.e., domain shifts), respectively: Original (no augmentation), Affine, Gamma, HSV, and RHM. We consider a DNN to be robust to a particular class of transformations if its performance does not degrade significantly, compared to unaltered test data, when those transformations are applied to the test set.  To investigate this we train multiple DNNs: each one is trained on a dataset with just one of five aforementioned augmentations. We then test each of these DNNs on a test set that has a matching augmentation. By training and testing with identical augmentations, we ensure that any performance degradations (compared to ``Original'' - the case of no augmentations) cannot arise due to a mismatch between the training augmentation and the target domain augmentation; in this case, performance degradation should only arise due to (i) inability of the model to become robust or (ii) loss of image information due to the augmentations (e.g., some augmentations compress the imagery).

The results of this experiment are presented on the diagonal (bolded) entries in \Cref{tab:training_and_testing_with_varying_domain_shifts}. As we see, the performance degradation for all domain shifts is relatively low, suggesting that the DNNs do achieve relative robustness when trained with a matching augmentation. We also applied each trained model to all of the other test datasets (i.e., that have different augmentations), and the results of this are presented in the un-bolded entries in \Cref{tab:training_and_testing_with_varying_domain_shifts}. As expected, substantial performance degradation is observed when applying the ``Original'' model to any of the augmented testing datasets. This confirms the importance of spectral augmentation of some kind when training DNNs with overhead imagery. Furthermore, when an augmentation is applied to the test set, we see that the best model for any testing dataset is the model that was trained with a matching augmentation. 

\begin{table}[h]
\centering
\caption{Performance in terms of intersection-over-union (IoU) of a U-Net model with a ResNet-50 encoder-decoder structure when trained and tested on data with different classes of augmentation. Bolded text indicates that the model was trained and tested on imagery with the same augmentation.}
\label{tab:training_and_testing_with_varying_domain_shifts}
\newcolumntype{C}{>{\centering\arraybackslash}X}
\begin{tabularx}{\linewidth}{C|CCCCC}
\toprule
\multirow[m]{2}{*}{\textbf{Training}} & \multicolumn{5}{c}{\textbf{Testing}} \\
% \textbf{Training} & \multicolumn{5}{c}{\textbf{Testing}} \\
& Original & Affine & Gamma & HSV & RHM \\
\midrule
Original & \textbf{0.746} & 0.367 & 0.496 & 0.524 & 0.547 \\
Affine & 0.739 & \textbf{0.725} & 0.728 & 0.669 & 0.646 \\
Gamma & 0.739 & 0.708 & \textbf{0.734} & 0.636 & 0.650 \\
HSV & 0.725 & 0.681 & 0.703 & \textbf{0.708} & 0.631 \\
RHM & 0.720 & 0.507 & 0.652 & 0.635 & \textbf{0.713} \\
\bottomrule
\end{tabularx}
\end{table}

Interestingly, we find that training with an augmentation that does not match the test set augmentation usually results in further performance degradation (i.e., compared to a matching augmentation strategy), and sometimes a substantial degradation. This has several implications.  First, these results suggest that spectral augmentations do not result in \textit{general} robustness to spectral domain shifts, so that the DNN learns to ignore many or most kinds of spectral shifts.  Instead, it appears that they confer robustness only for the class of domain shifts that were presented in training.  A corollary of this is that it is important to choose an augmentation strategy that does indeed emulate the domain shifts that can be expected in real-world imagery, and that failing to do so can result in substantial loss of otherwise recoverable performance.  

\begin{figure}[H]
\centering
\includegraphics[width=\linewidth]{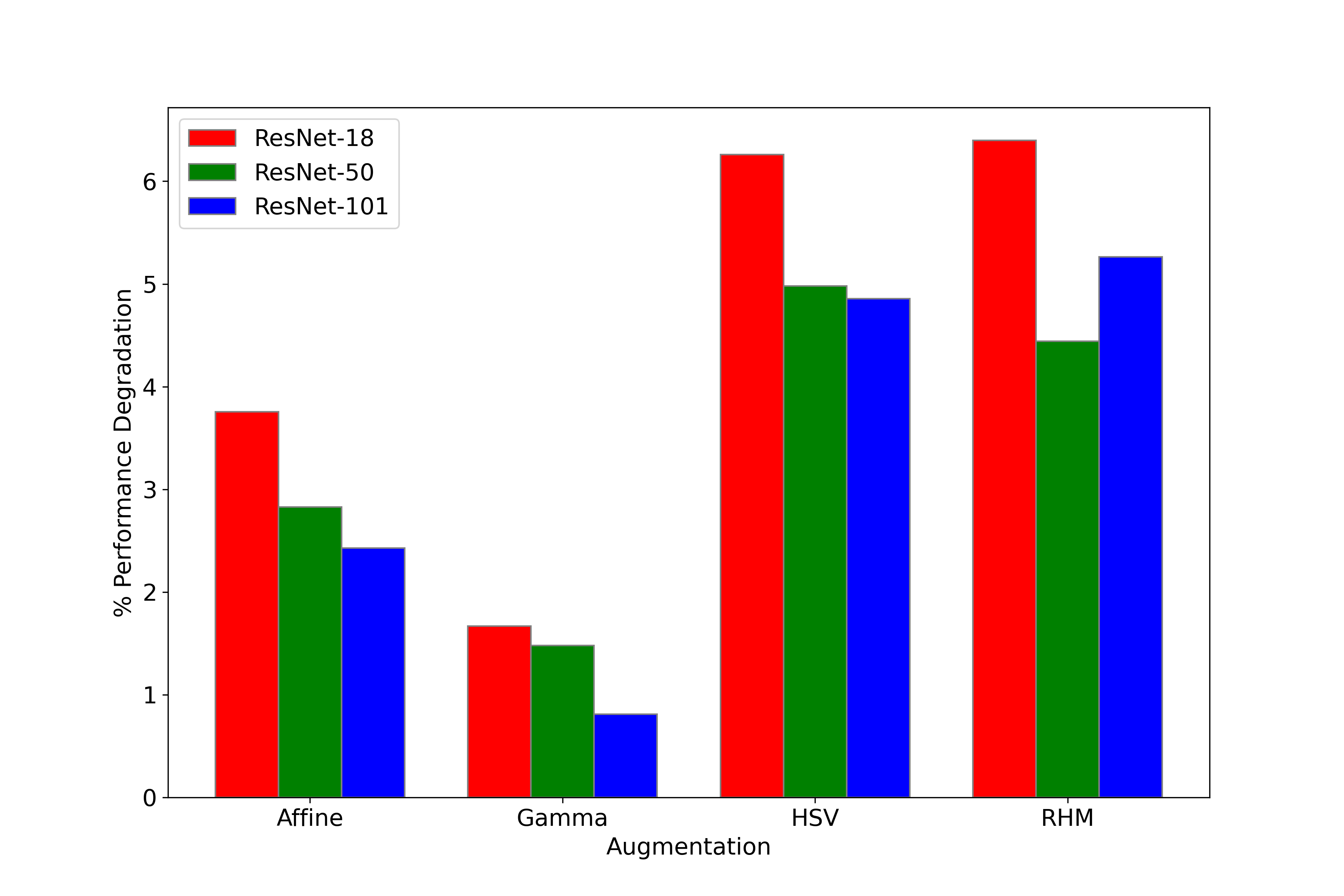}
\caption{Percent performance degradation as a function of model size and augmentation type. For each combination of model size and training augmentation type, we measure the percentage performance degradation when testing on the augmented test set compared to the un-augmented test set.}
\label{fig:does_augmentation_confer_invariance}
\end{figure}

We also investigate the extent to which DNN robustness depends upon the capacity of the model (e.g,. the number of free parameters it has). Therefore, we repeated our experiments using segmentation models with three different encoder sizes: ResNet-18, ResNet-50, and ResNet-101.  The results of this experiment are presented in \Cref{fig:does_augmentation_confer_invariance}, where we only report results when we train and test with the same augmentations. The results indicate that a larger model does seem to enable a greater level of invariance; except for RHM where ResNet-50 has slightly better performance than ResNet-101, the ResNet-18 and ResNet-101 models consistently perform the worst and best, respectively.

\section{Benchmark Results: One-to-One Domain Adaptation}
\label{sec:one_to_one}

In this section we compare RHM to other unsupervised domain adaptation methods when evaluated in a one-to-one scenario, i.e., we are given a single source domain, and we must maximize performance on a single (unlabeled) target domain \cite{tasar2020colormapgan,tasar2020semi2i}. Following the practice of recent work \cite{tasar2020colormapgan}, we treat the imagery over a single city as a single domain, and we randomly chose two pairs of cities (i.e., four total cities) for our one-to-one experiments. For each pair, we alternately trained on one of the two cities, and tested on the other. The only constraint on the selection of the city pairs is that they must contain one city from the Inria dataset, and one city from DeepGlobe (DG) dataset. These two datasets were produced by different groups and at different times, and therefore we reason they are more likely to exhibit domain shifts. All experiments were conducted with a U-Net model with a ResNet-50 encoder, as described in \Cref{sec:segmentation_model}. All results are reported in terms of intersection-over-union (IoU).  

\begin{table*}[ht]
\caption{Benchmark of domain adaptation methods for one-to-one using a ResNet-50 encoder. All results are reported in terms of intersection-over-union (IoU)}
%\begin{adjustwidth}{-\extralength}{0cm}
\label{tab:one_to_one}
% Feel free to change column widths using these definitions :)
%\newcolumntype{C}{>{\centering\arraybackslash}X}
\newcolumntype{s}{>{\centering\arraybackslash\hsize=.9\hsize}X}
\newcolumntype{C}{>{\centering\arraybackslash\hsize=1.25\hsize}X}
\begin{tabularx}{\linewidth}{CCsssss}%1.05\fulllength}{CCCCCCC}
\toprule
\textbf{Method} & \textbf{Type}	& \textbf{Vienna $\rightarrow$ Vegas} & \textbf{Vegas $\rightarrow$ Vienna} & \textbf{Tyrol-w $\rightarrow$ Shanghai} & \textbf{Shanghai $\rightarrow$ Tyrol-w} & \textbf{Average} \\
\midrule
Original & Naive & 0.342 & 0.503 & 0.416 & 0.398 & 0.415 \\
\midrule
Affine & \multirow{5}{*}{Augmentation} & 0.637 & \textbf{0.642} & 0.337 & 0.472 & 0.522 \\
Gamma & & 0.583 & 0.459 & 0.336 & 0.351 & 0.432 \\
HSV & & 0.600 & 0.565 & 0.250 & 0.516 & 0.483 \\
RHM (w/o res.) & & 0.627 & 0.584 & 0.398 & 0.557 & 0.542 \\
RHM (w/ res.) & & \textbf{0.646} & 0.597 & \textbf{0.433} & 0.534 & \textbf{0.553} \\
\midrule
Hist-Eq \cite{gonzalez2002digital} & \multirow{2}{*}{Standardization} & 0.500 & 0.531 & 0.260 & 0.505 & 0.449 \\
Gray-World \cite{buchsbaum1980spatial} & & 0.436 & 0.517 & 0.336 & 0.469 & 0.440 \\
\midrule
Hist-Match \cite{gonzalez2002digital} & \multirow{2}{*}{I2I Translation} & 0.558 & 0.527 & 0.317 & 0.501 & 0.476 \\
ColorMapGAN\cite{tasar2020colormapgan} & & 0.365 & 0.429 & 0.291 & \textbf{0.562} & 0.412 \\
\bottomrule
\end{tabularx}
%\end{adjustwidth}
\end{table*}

Descriptions of our baseline methods can be found in \Cref{sec:baseline_methods}.  As baselines, we included a variety of methods that are comparable in their simplicity and speed to RHM (e.g., HSV and Gamma augmentation, Gray-World standardization, etc.). We also included ColorMapGAN \cite{tasar2020colormapgan}, which is a more sophisticated approach which recently reported superior results to a large number of other state-of-the-art unsupervised domain adaptation methods when evaluated in the one-to-one scenario. As an ablation study, we also test RHM \textit{without} the entropy-based resampling step described in \Cref{subsec:entropy_based_resampling}.

The results of the benchmark experiments are presented in \Cref{tab:one_to_one}. RHM with resampling gives a slightly better IoU than RHM without resampling, while both RHM models perform substantially better than all other baselines (on average). RHM achieves the highest IoU on two of the four individual test cities (Vienna $\rightarrow$ Vegas and Tyrol-w $\rightarrow$ Shanghai). Although it does not achieve the highest performance on Vegas $\rightarrow$ Vienna or Shanghai $\rightarrow$ Tyrol-w, in both cases it is the second best performing model, and achieves very similar performance to the top-performing approach.

\section{Benchmark Results: Many-to-Many Domain Adaptation} 
\label{sec:many_to_many}
In this section, we compare RHM to other unsupervised domain adaptation methods when evaluated in a many-to-many scenario, i.e., we are given multiple source domains, and we must maximize performance on multiple (unlabeled) target domains\cite{tasar2020colormapgan,tasar2020semi2i}. In contrast to the one-to-one scenario, the many-to-many setting is more likely to reflect real-world testing conditions in which a model is trained on a large and diverse training set and then tested on multiple new collections of imagery (i.e., multiple target domains). For these experiments, we train each model on one of our two multi-city benchmark datasets (Inria and DG), and test on the other.

Because ColorMapGAN was designed specifically for the one-to-one task, and therefore may be at a disadvantage \cite{tasar2020colormapgan}, for these experiments we utilized an additional benchmark method, CycleGAN \cite{zhu2017unpaired}. CycleGAN recently achieved comparable performance to ColorMapGAN in \cite{tasar2020colormapgan}, while being better-suited for the many-to-many testing scenario. As in the previous section, we also test RHM with, and without, the entropy resampling step.

% UNCOMMENT NEXT TWO LINES IF LANDSCAPE IS PREFERRED
%\begin{landscape} 
%\begin{table}[H]
\begin{table*}[ht]
\caption{Benchmark of many-to-many domain adaptation methods for Inria $\rightarrow$ DeepGlobe using a ResNet-50 encoder. All results are reported in terms of intersection-over-union (IoU)}
\label{tab:inria_to_dg}
% Feel free to change column widths using these definitions :)
%\newcolumntype{C}{>{\centering\arraybackslash}X}
\newcolumntype{s}{>{\centering\arraybackslash\hsize=.9\hsize}X}
\newcolumntype{C}{>{\centering\arraybackslash\hsize=1.3\hsize}X}
%\begin{tabularx}{\linewidth}{CCCCCCCC}%1.5\fulllength}{CCCCCCCC}\
\begin{tabularx}{\linewidth}{CCssssss}
\toprule
\textbf{Method} & \textbf{Type} & \textbf{Khartoum} & \textbf{Paris} & \textbf{Shanghai} & \textbf{Vegas} & \textbf{Overall} & \textbf{City Average} \\
\midrule
Original & Naive & 0.270 & 0.365 & 0.467 & 0.684 & 0.533 & 0.447 \\
\midrule
Affine & \multirow{5}{*}{Augmentation} & 0.215 & 0.484 & 0.443 & 0.736 & 0.542 & 0.470 \\
Gamma & & 0.206 & 0.422 & 0.435 & 0.735 & 0.539 & 0.450 \\
HSV & & 0.159 & 0.427 & 0.391 & 0.740 & 0.519 & 0.429 \\
RHM (w/o res.) & & 0.244 & 0.525 & 0.494 & 0.746 & 0.583 & 0.502 \\
RHM (w/ res.) & & 0.260 & \textbf{0.556} & 0.518 & \textbf{0.746} & 0.590 & \textbf{0.520} \\
\midrule
Hist-Eq \cite{gonzalez2002digital} & \multirow{2}{*}{Standardization} & 0.166 & 0.316 & 0.367 & 0.663 & 0.477 & 0.378 \\
Gray-World \cite{buchsbaum1980spatial} & & 0.209 & 0.379 & 0.399 & 0.716 & 0.519 & 0.426 \\
\midrule
Hist-Match \cite{gonzalez2002digital} & \multirow{3}{*}{I2I Translation} & 0.144 & 0.282 & 0.279 & 0.734 & 0.432 & 0.360 \\
ColorMapGAN \cite{tasar2020colormapgan} & & 0.087 & 0.260 & 0.285 & 0.706 & 0.450 & 0.335 \\
CycleGAN \cite{zhu2017unpaired} & & \textbf{0.340} & 0.470 & \textbf{0.534} & 0.730 & \textbf{0.595} & 0.519 \\
\bottomrule
\end{tabularx}
\end{table*}

\begin{table*}[ht]
\caption{Benchmark of many-to-many domain adaptation methods for DeepGlobe $\rightarrow$ Inria using a ResNet-50 encoder. All results are reported in terms of intersection-over-union (IoU)}
\label{tab:dg_to_inria}
% Feel free to change column widths using these definitions :)
%\newcolumntype{C}{>{\centering\arraybackslash}X}
\newcolumntype{s}{>{\centering\arraybackslash\hsize=.82\hsize}X}
\newcolumntype{C}{>{\centering\arraybackslash\hsize=1.54\hsize}X}
\begin{tabularx}{\linewidth}{CCsssssss}%{1.5\fulllength}{CCCCCCCCC}
\toprule
\textbf{Method} & \textbf{Type} & \textbf{Austin} & \textbf{Chicago} & \textbf{Kitsap} & \textbf{West Tyrol} & \textbf{Vienna} & \textbf{Overall} & \textbf{City Average} \\
\midrule
Original & Naive & 0.396 & 0.310 & 0.491 & 0.472 & 0.592 & 0.424 & 0.452 \\
\midrule
Affine & \multirow{5}{*}{Augmentation} & 0.431 & 0.394 & 0.588 & 0.553 & 0.647 & 0.492 & 0.523 \\
Gamma & & 0.314 & 0.302 & 0.502 & 0.491 & 0.564 & 0.392 & 0.435 \\
HSV & & 0.493 & 0.377 & \textbf{0.605} & 0.559 & 0.635 & 0.494 & 0.534 \\
RHM (w/o res.) & & 0.562 & 0.576 & 0.579 & 0.649 & 0.664 & 0.613 & 0.606 \\
RHM (w/ res.) & & 0.574 & 0.573 & 0.601 & \textbf{0.658} & \textbf{0.671} & 0.618 & \textbf{0.615} \\
\midrule
Hist-Eq \cite{gonzalez2002digital} & \multirow{2}{*}{Standardization} & 0.560 & 0.566 & 0.568 & 0.629 & 0.634 & 0.597 & 0.591 \\
Gray-World \cite{buchsbaum1980spatial} & & 0.343 & 0.287 & 0.560 & 0.493 & 0.614 & 0.401 & 0.459 \\
\midrule
Hist-Match \cite{gonzalez2002digital} & \multirow{3}{*}{I2I Translation} & 0.544 & 0.568 & 0.577 & 0.566 & 0.663 & 0.599 & 0.584 \\
ColorMapGAN \cite{tasar2020colormapgan} & & 0.540 & 0.561 & 0.448 & 0.496 & 0.637 & 0.576 & 0.536 \\
CycleGAN \cite{zhu2017unpaired} & & \textbf{0.625} & \textbf{0.585} & 0.590 & 0.595 & 0.656 & \textbf{0.619} & 0.610 \\
\bottomrule
\end{tabularx}
\end{table*}
%\end{landscape}

Our many-to-many experimental results are reported in \Cref{tab:inria_to_dg} and \Cref{tab:dg_to_inria}, respectively. In each case, the IoU for each testing city is provided along with an ``Overall'' IoU (computed after aggregating all test city predictions) and a ``City Average'' (computed by averaging the IoUs of each test city). As with the one-to-one setting, we find that entropy-based resampling improves both the overall and per-city performance of RHM (in all nine cities, except for Chicago). Moreover, RHM outperforms CycleGAN on 5 of the 9 cities, and achieves better city average performance than all other baselines, while having a very similar overall IoU to CycleGAN in both training/testing directions. Notably, RHM gives comparable or better performance than CycleGAN without the need to train an auxiliary DNN (e.g., our CycleGAN typically required over three days to train on an NVIDIA Titan RTX), or tune many hyperparameters. Compared to other simple unsupervised domain adaptation approaches, RHM provides substantially better average performances on both benchmark test sets.

\section{Run-time Analysis}
In this section, we demonstrate that employing RHM during training only incurs modest computational costs compared to similar online augmentation approaches. Following the model and optimizer configurations outlined in \Cref{sec:segmentation_model}, we train a U-Net with a ResNet-18 encoder over a single epoch of the Inria dataset and benchmark the wall times of an average single training iteration for each baseline augmentation given in \Cref{tab:augmentations}, as well as RHM with entropy-based resampling. In \Cref{tab:walltime}, we report the increase in wall time compared to no augmentation used during training. 

\begin{table}[h] 
\captionsetup{justification=centering}
\caption{Increase in training time due to use of various online augmentation approaches with a ResNet-18 encoder. }
\label{tab:walltime}
\centering
\newcolumntype{C}{>{\centering\arraybackslash}X}
\begin{tabularx}{\linewidth}{CC}
\toprule
\textbf{Augmentation} & \textbf{\% Increase} \\
\midrule
Affine & 0.62 \\
Gamma & 1.25 \\
HSV & 4.08 \\
RHM (w/ res.) & 1.88 \\
\bottomrule
\end{tabularx}
\end{table}

We see that the use of RHM only results in a marginal increase in training time that is comparable to common baselines, and is in fact considerably faster than the widely-used HSV augmentation. We note that our implementation of RHM computes histograms on-the-fly during training, and does not reuse previously computed histograms in future iterations. At the cost of additional memory usage, it is possible to make RHM even faster by precomputing all histograms before training so that they are readily available for the matching process, but this is beyond the scope of our work.

\section{Conclusion}
In this work, we address the problem of unsupervised domain adaptation in overhead imagery. To do so, we model domain shifts caused by variations in imaging hardware, lighting conditions (e.g., due to time-of-day), or atmospheric conditions as non-linear pixel-wise transformations, and we show that DNNs can become largely robust to these types of transformations if they are provided with the appropriate training augmentation.  In general, however, we do not know the transformation between any two sets of imagery.  To overcome this problem, we propose randomized histogram matching (RHM), a simple real-time training data augmentation approach. We then conduct experiments with two large benchmark datasets for building segmentation and we find that RHM consistently yields comparable performance to recent state-of-the-art unsupervised domain adaptation approaches for overhead imagery, despite being substantially easier and faster to use.  RHM also offers substantially better performance than other comparably simple and widely-used unsupervised approaches for overhead imagery. This new approach to training augmentation has the ability to expand the efficacy of automated analysis of remote sensing data to more applications while reducing the burden of expensive labeled imagery from target domains.

\section*{Acknowledgments}
We would like to thank the Energy Initiative at Duke University for supporting this work.  

\bibliographystyle{IEEEtran}
\bibliography{references}

\newpage

%\section{Biography Section}
%If you have an EPS/PDF photo (graphicx package needed), extra braces are
% needed around the contents of the optional argument to biography to prevent
% the LaTeX parser from getting confused when it sees the complicated
% $\backslash${\tt{includegraphics}} command within an optional argument. (You can create
% your own custom macro containing the $\backslash${\tt{includegraphics}} command to make things
% simpler here.)
 
\vspace{11pt}

%\bf{If you include a photo:}
\vspace{-33pt}
\begin{IEEEbiography}[{\includegraphics[width=1in,height=1.25in,clip,keepaspectratio]{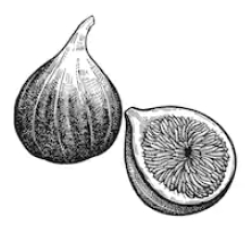}}]{Michael Shell}
Use $\backslash${\tt{begin\{IEEEbiography\}}} and then for the 1st argument use $\backslash${\tt{includegraphics}} to declare and link the author photo.
Use the author name as the 3rd argument followed by the biography text.
\end{IEEEbiography}

\vspace{11pt}

%\bf{If you will not include a photo:}
\vspace{-33pt}
\begin{IEEEbiographynophoto}{John Doe}
Use $\backslash${\tt{begin\{IEEEbiographynophoto\}}} and the author name as the argument followed by the biography text.
\end{IEEEbiographynophoto}

\vfill

\end{document}

%
% paper title
% Titles are generally capitalized except for words such as a, an, and, as,
% at, but, by, for, in, nor, of, on, or, the, to and up, which are usually
% not capitalized unless they are the first or last word of the title.
% Linebreaks \\ can be used within to get better formatting as desired.
% Do not put math or special symbols in the title.
\title{Supplementary Information: Randomized Histogram Matching}
%
%
% author names and IEEE memberships
% note positions of commas and nonbreaking spaces ( ~ ) LaTeX will not break
% a structure at a ~ so this keeps an author's name from being broken across
% two lines.
% use \thanks{} to gain access to the first footnote area
% a separate \thanks must be used for each paragraph as LaTeX2e's \thanks
% was not built to handle multiple paragraphs
%

\author{Can~Yaras,
        Kaleb~Kassaw,
        Bohao~Huang,
        Kyle~Bradbury,
        Jordan M.~Malof,~\IEEEmembership{Member,~IEEE}
        
\thanks{J.M. Malof is with the Department
of Electrical and Computer Engineering, Duke University, Durham,
NC, 27705 USA e-mail: (see http://www.michaelshell.org/contact.html).}% <-this % stops a space
\thanks{J. Doe and J. Doe are with Anonymous University.}% <-this % stops a space
\thanks{Manuscript received April 19, 2005; revised September 17, 2014.}}

% note the % following the last \IEEEmembership and also \thanks - 
% these prevent an unwanted space from occurring between the last author name
% and the end of the author line. i.e., if you had this:
% 
% \author{....lastname \thanks{...} \thanks{...} }
%                     ^------------^------------^----Do not want these spaces!
%
% a space would be appended to the last name and could cause every name on that
% line to be shifted left slightly. This is one of those "LaTeX things". For
% instance, "\textbf{A} \textbf{B}" will typeset as "A B" not "AB". To get
% "AB" then you have to do: "\textbf{A}\textbf{B}"
% \thanks is no different in this regard, so shield the last } of each \thanks
% that ends a line with a % and do not let a space in before the next \thanks.
% Spaces after \IEEEmembership other than the last one are OK (and needed) as
% you are supposed to have spaces between the names. For what it is worth,
% this is a minor point as most people would not even notice if the said evil
% space somehow managed to creep in.

% The paper headers
\markboth{Journal of \LaTeX\ Class Files,~Vol.~13, No.~9, September~2014}%
{Shell \MakeLowercase{\textit{et al.}}: Bare Demo of IEEEtran.cls for Journals}
% The only time the second header will appear is for the odd numbered pages
% after the title page when using the twoside option.
% 
% *** Note that you probably will NOT want to include the author's ***
% *** name in the headers of peer review papers.                   ***
% You can use \ifCLASSOPTIONpeerreview for conditional compilation here if
% you desire.

% If you want to put a publisher's ID mark on the page you can do it like
% this:
%\IEEEpubid{0000--0000/00\$00.00~\copyright~2014 IEEE}
% Remember, if you use this you must call \IEEEpubidadjcol in the second
% column for its text to clear the IEEEpubid mark.

% use for special paper notices
%\IEEEspecialpapernotice{(Invited Paper)}

% make the title area
\maketitle

% As a general rule, do not put math, special symbols or citations
% in the abstract or keywords.
\begin{abstract}
\textbf{To be completed.}  

\end{abstract}

% Note that keywords are not normally used for peerreview papers.
\begin{IEEEkeywords}
IEEEtran, journal, \LaTeX, paper, template.
\end{IEEEkeywords}

% For peer review papers, you can put extra information on the cover
% page as needed:
% \ifCLASSOPTIONpeerreview
% \begin{center} \bfseries EDICS Category: 3-BBND \end{center}
% \fi
%
% For peerreview papers, this IEEEtran command inserts a page break and
% creates the second title. It will be ignored for other modes.
\IEEEpeerreviewmaketitle

%%%%%%%%% BODY TEXT
\section{Additional Details: Does augmentation confer invariance to spectral domain shifts?}
\label{sec:does_augmentation_confer_invariance_supplement}

Here we provide additional details about Section 7 in the main paper where we investigated whether augmentation can confer invariance to spectral domain shifts. 
We begin in subsection \ref{sec:does_augmentation_confer_invariance_supplement_formulation} by providing a more detailed theoretical formulation of the problem, and use it to provide additional justification for the experimental design that we employed in Fig. 4 and Section 7 of the main paper.  In subsection \ref{sec:does_augmentation_confer_invariance_supplement} we begin by presenting a theoretical formulation of model invariance to spectral domain shifts.  This theoretical exposition is helpful to justify and interpret the experimental design we employed in Section 7 of the main paper, as well as to support the discussion in sub-section \ref{sec:does_augmentation_confer_invariance_supplement_information_loss}, where we discuss one limitation of our experimental design due to image information loss caused by our image augmentation methods.      

\subsection{Problem formulation and experimental design}
\label{sec:does_augmentation_confer_invariance_supplement_formulation}

Let $X^{s} \in \mathbb{R}^{NxNx3}$ and $Y^{s} \in \mathbb{R}^{NxN}$ be two random variables corresponding to source domain images and pixel-wise class labels, respectively, with joint probability distribution given by $Pr(X^{s},Y^{s})$.  Then we hypothesize that our target domain is given by a random variable $X^{t} \in \mathbb{R}^{NxNx3}$, and is related to the source domain by
\begin{equation}
\label{eq:domain_shift_function}
X^{t}=f_{\Phi}(X^{s}), 
\end{equation}
where $\Phi$ is a random variable that parameterizes a class of spectral image transformations (as described in Eq. (1) in the main paper) so that $f_{\Phi}: R^{3} \xrightarrow{} R^{3}$ is a non-linear function applied independently to each pixel in the input image.  For example, in our experiments in Section 7 of the main paper we considered the following classes of pixel-wise transformations: Affine, Gamma, HSV, and the space of functions induced by Randomized Histogram Matching (RHM).    

It will be useful in the subsequent discussion to define a precise performance metric for our DNN segmentation models.  Let $\mathcal{R}(m(X),Y)$ be a reward function (e.g., intersection-over-union) indicating the relative accuracy of predictions made by a trained DNN model, $m(X)$, as compared to the ground truth labels, $Y$.  Then we define the expected prediction reward, $r$, of the model to be given by  
\begin{equation}
    \label{eq:model_error_theoretical}
    r(m,X,Y) \triangleq E_{X,Y,\Phi}[\mathcal{R}(m(X),Y)] 
\end{equation}
This will serve in the subsequent discussion as our primary performance metric for a trained DNN.  

We can now proceed with a definition of spectral invariance.  A major hypothesis in our work was that spectral domain shifts might be addressed simply by making a conventional DNN invariant to spectral domains shifts.  Given these particular goals, we want to say that a model is invariant if it satisfies the following two conditions.  (i) First, it does not change its output predictions in the presence of a spectral domain shift (i.e., $f_{\Phi}$) of the input image.  For our experiments in the main paper, we adopted a more convenient approximation to this criterion; we require instead that the model's expected prediction error does not vary with respect to spectral domain shifts.  Mathematically, we require that $r(m,f_{\Phi=\phi}(X),Y)$ does not vary with respect to $\phi$ (i.e., particular realizations of the random variable $\Phi$).

As a second condition (ii), we require that a model should achieve criterion (i) without degradations in performance with respect to a conventional state-of-the-art DNN, so that making the model invariant does not come at the cost of worse overall performance.  Let $m^{0}$ be a conventional DNN model (e.g., U-net trained on source domain training data).  Now we can combine conditions (i) and (ii) into a single condition for invariance. In particular, we say that a given model, $m^{*}$ is invariant to spectral domain shift if it satisfies the following condition: 
\begin{equation}
    \label{eq:invariance_condition_theoretical}
    \begin{split}
    r(m^{*},f_{\Phi}(X^{s}),Y^{s})-r(m^{0},X^{s},Y^{s}) \approx 0 \\
    \end{split}
\end{equation}
The first term above is the expected error of $m^{*}$ over the target domain, which includes all forms of spectral shift.  The second term is the performance of a conventional DNN on the source domain.  Therefore this expression requires that performance remains consistent with a conventional DNN over all potential spectral transformations, thereby satisfying criteria (i) and (ii).    

In Fig. 4 and Section 7 of the main paper we employed the criterion in Eq. \ref{eq:invariance_condition_theoretical} to argue that training DNN models with spectral augmentation made the DNNs approximately invariant to that same augmentation (e.g., Affine, HSV, Gamma transforms).  However, we observed that none of the models became \textit{perfectly} invariant by this criteria, although they were close.  However the model trained with RHM did not generalize well to the test data that was augmented RHM augmentations, so that Eq. \ref{eq:invariance_condition_theoretical} was not well satisfied.  In the next sub-section we will discuss why we believe the criterion in Eq. \ref{eq:invariance_condition_theoretical} is unlikely to be satisifed perfectly by any augmentation, and in some cases it is not possible for any model to even approximately satisfy it, which we believe is the case for RHM.   

\subsection{Image information loss from augmentation}
\label{sec:does_augmentation_confer_invariance_supplement_information_loss}

One implicit assumption of condition (i) underpinning Eq. \ref{eq:invariance_condition_theoretical} is that the target classes are equally discriminable before and \textit{after} applying the spectral transformation $f_{\Phi}$ to the input imagery.  If this is \textit{not} true then it is not generally possible for a recognition model to make the exact same predictions, or the same accuracy, before and after the transformation is applied.  If true, this would also imply that the condition in Eq. \ref{eq:invariance_condition_theoretical} is impossible to satisfy (in general), because $m^{*}$ is evaluated on the shifted imagery (i.e., $f_{\Phi}(X^{s})$), while $m^{0}$ is only evaluated on the original un-augmented imagery, which is (by assumption) is easier.  

We assert that the discriminability of the target classes does tend to decrease after applying the spectral transformations considered in our experiments.  This  can arise if $f_{\Phi}$ maps many distinct pixel intensities to the same values, reducing the image information content, and potentially also destroying information that is useful for recognizing the target classes.  Here we provide evidence that the augmentations employed in our experiments do indeed reduce the information content of the imagery, and sometimes substantially.   

\begin{figure}
\centering
\includegraphics[scale=0.55]{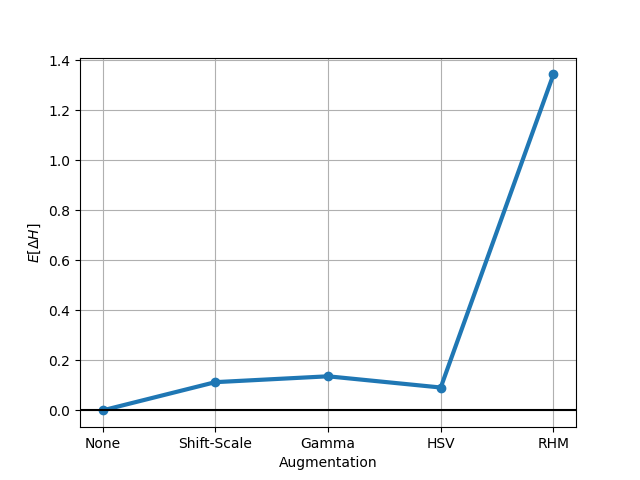}
\caption{The estimated reduction in image entropy caused by applying each of the different types of augmentation, respectively, to our Inria testing imagery. Larger values of $E[\Delta H]$ correspond to greater reductions in entropy, and therefore greater levels of image information reduction.}
\label{fig:expected_entropy_change}
\end{figure}

One method of evaluating the information content of a signal is by measuring its entropy, where reductions in entropy correspond to information loss in the signal.  Here we will estimate the expected change in entropy, $E_{\Phi,D^s}[\Delta H]$, for a particular type of augmentation from our experiments in Section 7 in the main paper.  The distribution for $\Phi$ for each type of augmentation is provided in Table 2 in the main paper.  Our sample estimator for $E_{\Phi,D^s}[\Delta H]$ is given by
\begin{equation}
    E[\Delta H] = \sum_{(x^{s},x^{t}) \in D} \sum_{i=1}^{3}  [H(x^{s}) - H(x^{t})]
\end{equation}
Here $x^{s}$ and $x^{t}$ refer to the original un-augmented source domain imagery and its augmented version, respectively, using a randomly sampled augmentation parameter.  The set $D$ refers to the full set of these image pairs, which are identical to the test set employed in our experiments in Section 7 of the main paper.    

The results of this experiment are presented in Fig. \ref{fig:expected_entropy_change}.  Note that positive values correspond to reductions in entropy, indicating image information loss.  Therefore we see that all of the augmentations employed in our experiments resulted in some reduction in image entropy.  Notably, we see that RHM leads to the greatest reduction in entropy, which is roughly proportional to its poor invariance results in Fig. 4 in the main paper.

\begin{figure*}
\centering
\includegraphics[scale=0.65]{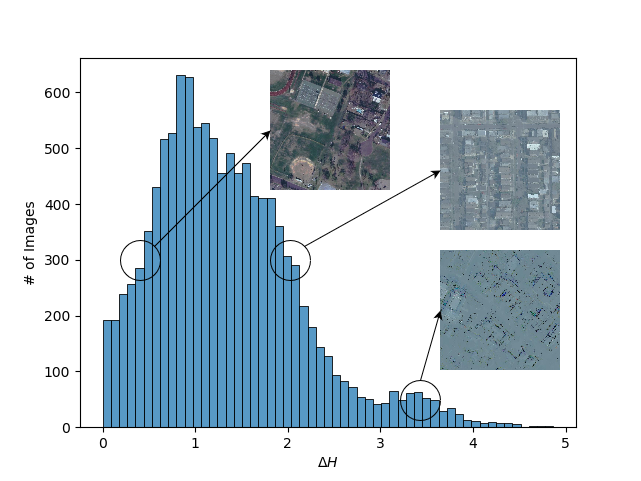}
\caption{A histogram of the change in entropy for each overhead image in our Inria testing dataset that was caused by Randomized Histogram Matching (RHM) approach. A positive $\Delta H$ value corresponds to a reduction in entropy, and therefore loss of image information.  We present examples of imagery corresponding to different values of $\Delta H$, where the increasing loss of information is apparent in the saturation of the pixel intensities in the images}
\label{fig:rhm_augmentation_histogram}
\end{figure*}